\documentclass[utf8]{FrontiersinHarvard_nofrontiers}

\usepackage{url,microtype,subcaption}
\usepackage[colorlinks,allcolors=black]{hyperref}
\usepackage[onehalfspacing]{setspace}
\usepackage{tabularx}

\def\keyFont{\fontsize{8}{11}\helveticabold }
\def\firstAuthorLast{Sabelhaus {et~al.}} 
\def\Authors{Andrew P. Sabelhaus\,$^{1,2*}$, Rohan Mehta\,$^{2}$, Anthony T. Wertz\,$^{2,3}$, and Carmel Majidi\,$^{2,3}$}
\def\Address{$^{1}$Soft Robotics Control Lab, Department of Mechanical Engineering, Boston University, Boston, MA, United States \\
$^{2}$Soft Machines Lab, Department of Mechanical Engineering, Carnegie Mellon University, Pittsburgh, PA, United States  \\
$^{3}$Robotics Institute, Carnegie Mellon University, Pittsburgh, PA, United States  }

\def\corrAuthor{Corresponding Author}
\def\corrEmail{asabelha@bu.edu}

\begin{document}
\onecolumn
\firstpage{1}

\title[In-Situ Sensing and Dynamics Predictions]{In-Situ Sensing and Dynamics Predictions for Electrothermally-Actuated Soft Robot Limbs} 

\author[\firstAuthorLast ]{\Authors} 

\address{} 

\correspondance{Andrew P. Sabelhaus, } 

\extraAuth{}

\maketitle

\begin{abstract}

\section{}
Untethered soft robots that locomote using electrothermally-responsive materials like shape memory alloy (SMA) face challenging design constraints for sensing actuator states.
At the same time, modeling of actuator behaviors faces steep challenges, even with available sensor data, due to complex electrical-thermal-mechanical interactions and hysteresis.
This article proposes a framework for in-situ sensing and dynamics modeling of actuator states, particularly temperature of SMA wires, which is used to predict robot motions.
A planar soft limb is developed, actuated by a pair of SMA coils, that includes compact and robust sensors for temperature and angular deflection.
Data from these sensors are used to train a neural network based on the long short-term memory (LSTM) architecture to model both unidirectional (single SMA) and bidirectional (both SMAs) motion.
Predictions from the model demonstrate that data from the temperature sensor, combined with control inputs, allow for dynamics predictions over extraordinarily long open-loop timescales (10 minutes) with little drift.
Prediction errors are on the order of the soft deflection sensor's accuracy.
This architecture allows for compact designs of electrothermally-actuated soft robots that include sensing sufficient for motion predictions, helping to bring these robots into practical application.

\tiny
 \keyFont{ \section{Keywords:} soft robot control, soft robot actuation, soft robot sensing, soft robot dynamics, soft robot modeling, machine learning, sensor design, shape memory alloy, artificial muscle} 
\end{abstract}

\section{Introduction}

Some of the most compact and easy-to-design artificial muscles for soft robots are based on electrothermal actuation (\cite{rich_untethered_2018}).
These actuators are typically composed of a thermally-responsive shape changing material that can be electrically stimulated with Joule heating.
Examples include coiled polymers that contract when heated (\cite{haines2014artificial,Pawlowski2019}), actuators containing shape memory alloy (SMA) wire (\cite{Rodrigue2017}), and liquid crystal elastomers (\cite{yuan_3d_2017,he_electrically_2019,ford_multifunctional_2019,kent_soft_2020}).

However, these easy-to-use designs sacrifice the easy-to-use models of other soft actuators (e.g., pneumatics) due to the their internal actuator dynamics alongside time-dependent hysteretic behaviors (\cite{xiang_design_2017,ge_preisach-model-based_2021}). 
In particular, SMA actuators with popular materials like Nitinol have dynamics that are dictated by complex interactions at the microscopic scale.
Hysteresis arises in residual stress, strain, material phase, and electrical resistance (\cite{zakerzadeh_modeling_2011}), so first-principles models necessarily include state variables that can only be sensed indirectly (\cite{Majima2001}).
Therefore, most modeling of SMA and other electrothermal actuators occurs in artificial test setups with extensive sensing capabilities (\cite{zakerzadeh_modeling_2011,lee_precise_2013,ge_preisach-model-based_2021}).

As a result of these modeling difficulties, as well as significant weight and volume restrictions for sensing and control hardware, there has yet to be an untethered electrothermally-actuated soft robot with embedded sensors and associated modeling that capture actuator states (\cite{rich_untethered_2018}).
In fact, most SMA-based soft walking robots often focus on designs that do not include embedded sensors and operate entirely through open-loop control or dependency on external hardware (\cite{Huang2018,Huang2019,patterson_untethered_2020,boothby_untethered_2021,thomas_untethered_2021,rehan_towards_2021,meng_mechanically_2020,almubarak_kryptojelly_2020,kim_shape_2021}).
In turn, modeling of these untethered robots, when it has been done, does not include time-dependent actuator states (\cite{huang_dynamic_2020,Goldberg2019}).

\subsection{Study Overview}

\begin{figure}[h!]
    \centering
    \includegraphics[width=0.5\textwidth]{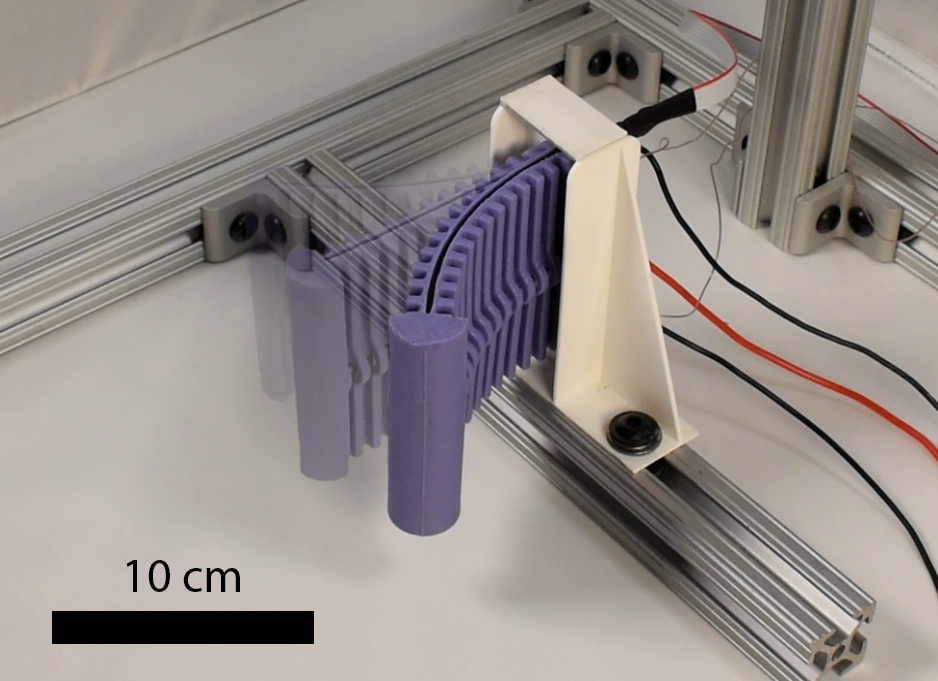}
    \caption{This article develops a framework for sensing and dynamics modeling of a soft, planar, electrothermally-actuated shape memory alloy (SMA) robot limb. Our model can predict the limb's deflection angle over long periods of time using combinations of power input and in-situ temperature measurements.}
    \label{fig:limb_motion}
\end{figure}

This work contributes a framework to bridge the gap between sensing capabilities and accurate dynamics predictions for a soft electrothermally actuated robot limb. 
The limb is actuated using SMA coils that are activated through electrical current and Joule heating.  
We hypothesize that temperature is a reasonable surrogate for full thermal actuator state due to the prior successes of \cite{ge_preisach-model-based_2021} and \cite{wertz_trajectory_2022} in tracking the behavior of SMA-based actuators.
Our framework includes a fabrication method to securely and robustly attach a compact temperature sensor to a highly-dynamic SMA wire, alongside a neural-network based model that captures hysteresis in sensor data versus limb pose.
Referring to Fig. \ref{fig:limb_motion}, the prototype limb is actuated by an antagonistic pair of SMA wire coils, and contains both the temperature sensors and a dielectric elastomer bend sensor embedded within a soft silicone rubber.
We use pulse-width-modulation (PWM) to apply electrical power.

After detailing sensor construction, we present a neural network architecture that can predict limb deflections based on different combinations of PWM and temperature data.
We collect a large dataset of temperature, deflection, and PWM.
Then we train our model and validate the model using open-loop simulations (Fig. \ref{fig:limb_motion}(C)).
For comparison, we perform a least-squares fit on the same data, as was used in \cite{wertz_trajectory_2022}, and show that our neural network makes predictions possible using PWM duty cycles only.
Our validation shows close alignment with hardware data when measurements from all of the sensors are included.

Specifically, this article makes the following contributions:
\begin{enumerate}
    \item A sensor design approach for compactly incorporating temperature measurements into SMA-based soft robot limbs,
    \item A proof-of-concept approach for learning dynamics models of soft electrothermally-actuated robot limbs using this sensor data,
    \item Open-loop rollout simulations of the robot limb's angular deflections, showing low error ({RMSE = 5.350$^\circ$}) and no significant drift over very long horizons (10 minutes or more).
\end{enumerate}

Combined, our results demonstrate that this type of robot limb architecture can be constructed to achieve relatively high-fidelity predictions of motion with only in-situ sensing. 
Results also demonstrate noticeable improvements of predictions under bidirectional bending, with two active SMAs, in comparison to unidirectional bending with only a single SMA, producing design implications.

Though our experiment employs shape memory alloy wires as electrothermal actuators, our modeling approach does not make any assumptions about the underlying actuator technology: we use only electrical power and temperature measurements.
We therefore expect these results to generalize to other electrothermal actuators exhibiting the same hysteretic characteristics, e.g. liquid crystal elastomer (LCE) based actuators with embedded liquid metal conductors for electrically-controlled Joule heating (\cite{ford_multifunctional_2019,kent_soft_2020,kotikian_innervated_2021}).
In this respect, the approach here has the potential to serve as a framework for modeling a wider range of compact soft robot limb architectures that incorporate embedded sensing.

\subsection{Modeling and Neural Network Background}

The analysis of our framework focuses on the \textit{in situ} sensing and modeling of our limb (i.e. no external equipment), analyzing the effects of different sensor data as well as different motions on learned predictions.
In contrast, existing state-of-the-art hysteresis modeling of electrothermal actuators focuses primarily on a single type of motion with a fixed set of external sensors (\cite{xiang_design_2017,ge_preisach-model-based_2021,luong_long_2021}).
We train networks both with and without that data and quantify the prediction quality that is possible given different sensors.
This work improves on the approach from \cite{sabelhaus2021gaussian}, which demonstrated the need for explicitly addressing hysteresis.

Prior work in soft and electrothermally-actuated robot motion has generally eschewed extensive dynamics modeling in favor of feedback control, intended to compensate for modeling errors.
Feedback has been used for soft electrothermal and shape-memory actuators, but when performed with embedded sensing only, has been limited to model-free approaches (\cite{ma_position_2004,Elahinia2002,kent_soft_2020,kotikian_innervated_2021}).
However, many benefits to modeling exist in the context of open-loop motion generation, including a-priori trajectory optimization \cite{wertz_trajectory_2022}.

For our neural network model, we choose the long short-term memory (LSTM) architecture to account for hysteresis in predictions of robot deflection based on actuator state.
There has been extensive successful prior work that uses LSTM architectures for soft robot sensing (\cite{thuruthel_soft_2019,kang_learning-based_2020,han_use_2018}), modeling (\cite{zhao2021shape}), and control (\cite{luong_long_2021}), as well as many other options for machine learning applied to soft robots (\cite{cheng_design_2019}).

As a result, this article does not seek to develop a new neural network architecture for SMA-based soft robots, nor compare performance among possible architectures to determined the highest-performance predictions.
Instead, we show proof-of-concept, demonstrating that it is indeed possible to combine sensing and dynamics predictions in a compact and robust design framework.
We anticipate that future comparisons of available architectures, including e.g. hierarchical networks (\cite{han_use_2018}), would improve our framework's performance.

\section{Materials and Methods}

\subsection{Hardware Platform and In-Situ Sensor Design}

Our proposed framework consists of a sensor design approach and a machine-learned predictor that uses the sensor's data.
For the sensor itself, this section details the design procedure and its application to a specific prototype used in the remainder of the article.

\subsubsection{Robot Platform Overview}

\begin{figure}[h!]
    \centering
    \includegraphics[width=0.5\textwidth]{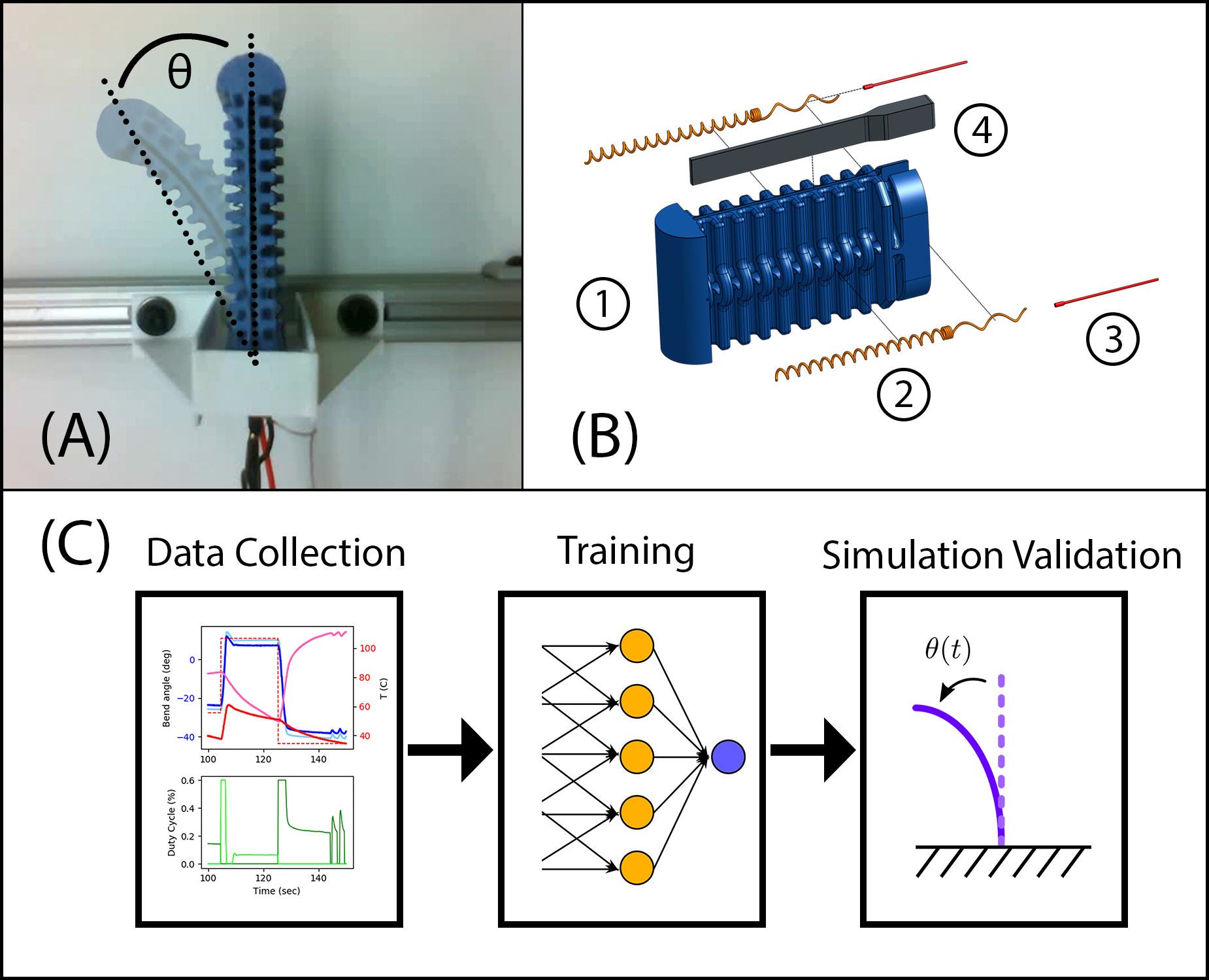}
    \caption{Overview of our in-situ sensor and learning procedure. (A) The prototype soft limb has a single degree-of-freedom, $\theta$, for planar motion. (B) The limb is cast from bulk silicone (B1), and is actuated by two shape-memory alloy (SMA) coils (B2) contained in a ridge along both sides of the limb. Our sensors include thermocouples attached to each SMA (B3) and a soft bending sensor (B4) for deflection and temperature measurements. Full assembly is detailed below. (C) Our framework consists of data collection of the limb undergoing various motions, used to train our LSTM neural network, with the outputs validated in simulated rollouts of deflection angle.}
    \label{fig:limb_overview}
\end{figure}

We consider the single-segment soft robot limb in Fig. \ref{fig:limb_motion} and Fig. \ref{fig:limb_overview}, derived from prior work on a legged soft robot (\cite{patterson_untethered_2020}). 
The limb is designed for planar motions only, reducing the dimensionality of the state space in order to test algorithmic approaches to modeling and control.
An earlier version of this limb was used in \cite{wertz_trajectory_2022}, and this article extends and details the fabrication procedure for the \textit{in situ} embedded temperature sensor.

The limb itself consists of a bulk silicone body (Smooth-On Smooth-Sil 945) embedded with actuators and sensors (Fig. \ref{fig:limb_overview}B).
Two SMA coil actuators extend along the length of each side of the limb (Fig. \ref{fig:limb_overview}(B1)) and are composed of nickel-titantium alloy (Dynalloy Flexinol, 0.020'' wire diameter).
A soft, rubbery capacitive bend sensor (Bendlabs, Inc.) is inserted into a groove in the limb (Fig. \ref{fig:limb_overview}(B4)).

A temperature sensor is attached to each SMA coil actuator using our proposed method.
The primary challenge in affixing a sensor to an SMA coil is the stress and strain induced during motion: a rigid thermal interface (such as a thermally conductive glue) can easily fracture.
As a result, though there are a variety of options for off-the-shelf temperature sensing, our approach uses a thermocouple due to its compact size.
The specific design here uses 30 AWG type K thermocouples (Omega Engineering).  
Because of its small size, only a small amount of adhesive is required to attach the thermocouple to the SMA coil, thereby reducing the region over which stress concentrations and interfacial failure can occur.

\subsubsection{Sensor Fabrication and Prototype}

\begin{figure}[h!]
    \centering
    \includegraphics[width=0.8\textwidth]{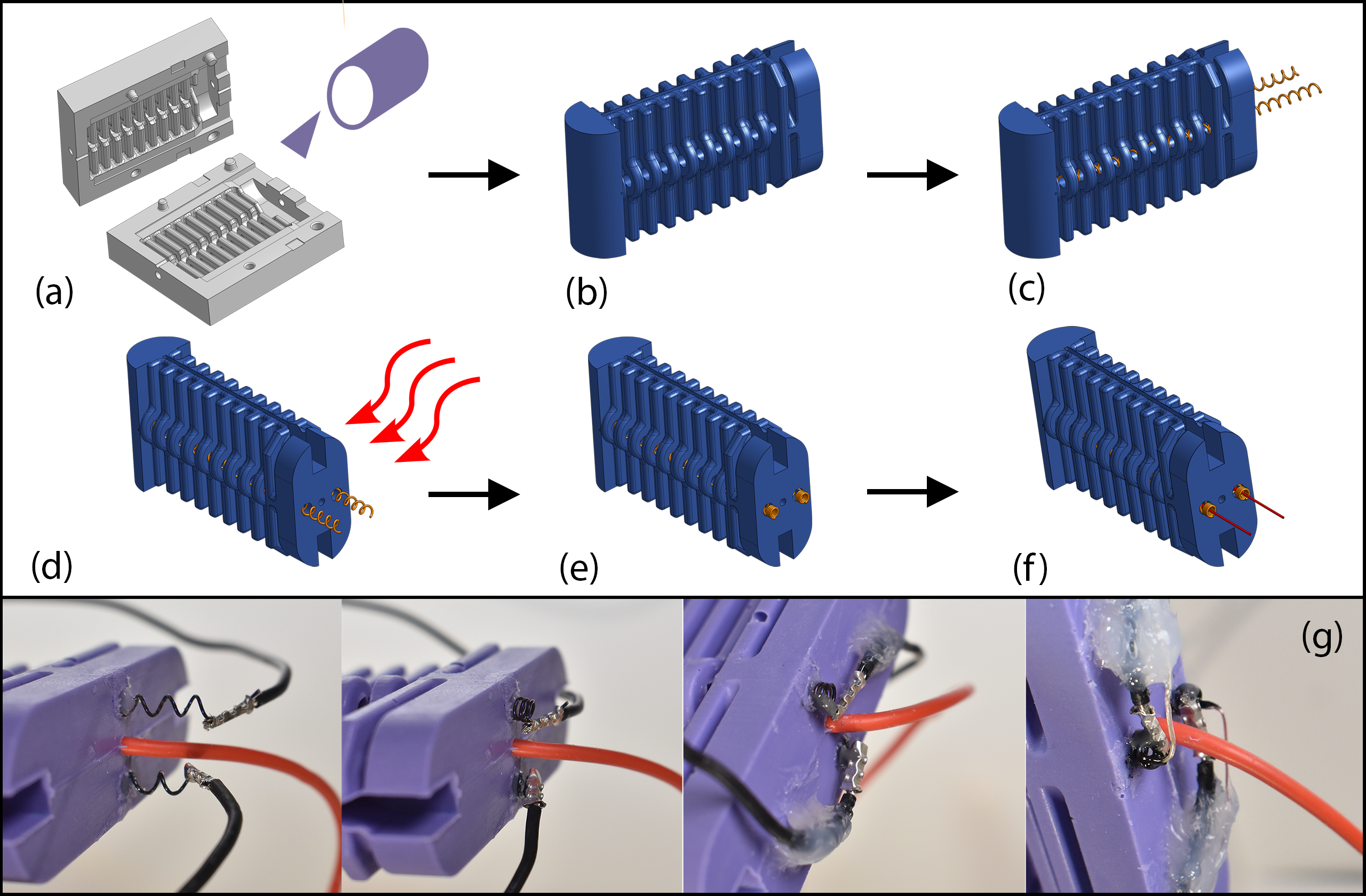}
    \caption{Fabrication procedure for our SMA-actuated soft limb with temperature sensing. (a) The limb is cast from bulk silicone using a 3D printed mold, then (b) released to form the main body. (c) SMA wires are inserted and sealed in place with adhesive, leaving additional turns of wire extending out the rear. (d) Heat is applied to the rear of the limb, so that (e) the extra turns of wire fully actuate. (f) Thermocouples are then adhered into the contracted coils with thermally-conductive epoxy, and experience no further mechanical stress during operation. (g) Fabrication in hardware follows this procedure.}
    \label{fig:limb_fab}
\end{figure}

The fabrication procedure for our limb and its sensors is shown in Fig. \ref{fig:limb_fab}.
First, as with our prior work in \cite{patterson_untethered_2020}, the limb is cast using a 3D-printed mold (Objet 30, Verowhite) and released after curing.
Next, the two SMA wire coils are inserted into the limb, crimped to electrical connections at both ends, and fixed into place at the tip and base using adhesive (Smooth-On Silpoxy).
Additional turns of wire are left exposed at the base (rear) of the limb.
We cut the wire to leave approximately 2-3 turns exposed.

Our key sensor fabrication step is to then heat the exposed turns of SMA coil until they fully actuate (Fig. \ref{fig:limb_fab}(g)).
Thanks to the shape memory effect, these turns of coiled wire will remain in their retracted state once they cool to room temperature and for all future thermal cycling, and crucially, no additional deflection will then occur during use.
We adhere the thermocouple tip to the retracted coils using thermally-conductive epoxy (MG 8329TCF), and once cured, apply additional soft adhesive to secure the thermocouple sheath to the robot body.

Finally, the sensors and SMAs were connected to a similar circuit as in \cite{patterson_untethered_2020} for sensing and actuation.
The thermocouple leads were connected to a digital thermocouple amplifier (MAX31855), and the SMAs to an N-channel power MOSFET each.
The thermocouple amplifier and soft capacitive bend sensor communicated digitally with a microcontroller.
The microcontroller actuated the MOSFETS using a pulse-width modulation (PWM) signal, with a nominal voltage of 7V from a benchtop power supply.
Though our test setup is tethered, these types of circuitry and power sources have previously been integrated into untethered robots (\cite{patterson_untethered_2020,Huang2019}).

The procedure presented here is tested with high-deflection electrothermal actuators (SMA coils) to which epoxy has a high bond strength.
However, we believe that the concept is generalizable to both straight-wire SMAs and other shape memory materials.
If a design were to leave an exposed portion of the actuator in a location where no stress occurs during use, and it were to actuate a single time, then that portion would remain actuated and not experience further stress.
Any appropriate adhesive may then be used to bond a thermocouple.

\subsection{Robot Modeling Setup and Data Collection}

Our proposed framework uses data from these sensors and inputs to train a neural network that predicts robot pose (bending angle).
Electrothermally-actuated robot limbs, such as that in this article, have internal state corresponding to both actuator dynamics and body dynamics.
Actuator dynamics are principally governed by temperature and the temperature response given a particular amount of electrical power input.
Body dynamics arise from applied actuator stress.

In lieu of first-principles models that demand access to many internal states, we employ available sensor data as the inputs to our predictor.
We consider four different usage scenarios: actuation with either one SMA (unidirectional) or both antagonistic SMAs (bidirectional), and with or without temperature data.
This allows us to determine the performance and applicability of our temperature sensor. 
The sensor input vector for the dynamics model of our soft electrothermal SMA-powered limb is denoted $v_{k}(t) \in \mathbb{R}^n$, where $n$ varies with our inclusion of either of the two actuators and/or temperature measurements, and $k$ denotes the particular combination of inputs:

\begin{align}
v_1(t) & = \begin{bmatrix} 
u_A(t)
\end{bmatrix} \label{eqn:v1}\\
v_2(t) & = \begin{bmatrix} 
u_A(t) & T_A(t)
\end{bmatrix} \label{eqn:v2} \\
v_3(t) & = \begin{bmatrix} 
u_A(t) & u_B(t)
\end{bmatrix} \label{eqn:v3} \\
v_4(t) & = \begin{bmatrix} 
u_A(t) & T_A(t) & u_B(t) & T_B(t)
\end{bmatrix}, \label{eqn:v4}
\end{align}

\noindent where $u_j$ and $T_j$ refer to the PWM duty cycle input and measured temperature of either SMA wire, $j \in [A,B]$. 

To generate the data for each of the above input spaces, two series of experiments were performed:

\begin{enumerate}
    \item One single SMA was actuated, causing unidirectional counterclockwise bending (+$\theta$ direction).
    \item Both SMAs were used to achieve bidirectional bending (both +$\theta$ and -$\theta$ directions).
\end{enumerate}

\begin{figure}[h!]
    \centering
    \includegraphics[width=0.5\textwidth]{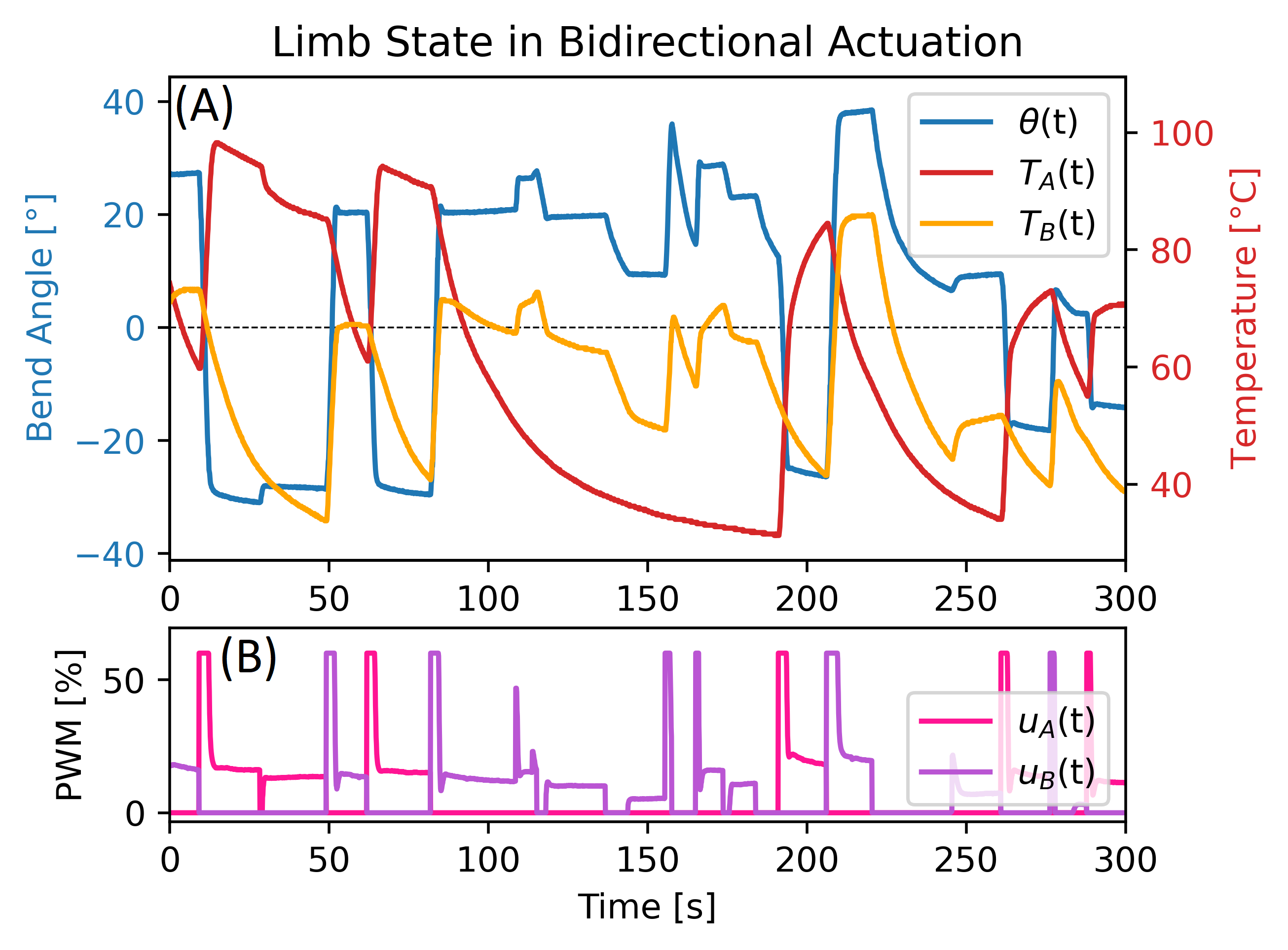}
    \caption{Data collection test example, bidirectional. (A) Temperature of each SMA in the limb during test alongside corresponding limb angle. (B) PWM input into each SMA during test at corresponding timepoints.}
    \label{fig:inputspace}
\end{figure}

To collect this data in a realistic application, we applied a simple controller to the limb to hold different setpoint angles as a form of ``motor babbling.''
Setpoints of $\bar \theta$ were sampled from a uniform distribution of $[0^\circ, 45^\circ]$ for $v_{1,2}$ and $[-45^\circ, 45^\circ]$ for $v_{3,4}$, with hold times also sampled from a uniform distribution of $[1, 30]$ sec. 
We used PI control for regulation, of the form $u = K_P e + \sum_t e \Delta t$, where $e = (\bar \theta - \theta)$.
Constants were tuned by hand to be $K_p = 0.06$ and $K_I = 1e^{-5}$.
The result was then saturated to $u \in [-1, 1]$ and mapped to PWM duty cycles as $u^+ \mapsto u_A$, $u^- \mapsto -u_B$.
This procedure was performed for approximately 6 hours of data, though only 30 minutes was used for training.
An example of states and inputs recorded during a test is shown in Fig. (\ref{fig:inputspace}).

\subsection{Neural Network Architecture}

The framework proposed in this article consists of a neural network based on long short-term memory (LSTM) to predict the bending angle of a soft limb, tested both in one-step-ahead predictions as well as open-loop rollouts.

\subsubsection{Long Short-Term Memory (LSTM) Networks}

LSTMs are a recurrent neural network (RNN) commonly applied to time series applications, due to their capability of learning order-based dependence (\cite{hochreiter1997long}).
Recursive Neural Networks (RNNs), while able to account for a short window of relevant information, typically have difficulty propagating information across long time lags.
This is referred to as the vanishing gradient problem, where error recurred through the hidden layers of an RNN can decay to zero or grow unbounded.
In contrast, the LSTM architecture ensures constant error flow.
This allows for complex information -- such as the hysteresis in stress, strain, etc. in soft electrothermal actuator dynamics -- to be carried through over a long period of time-dependence.

\begin{figure}[h!]
    \centering
    \includegraphics[width=0.5\textwidth]{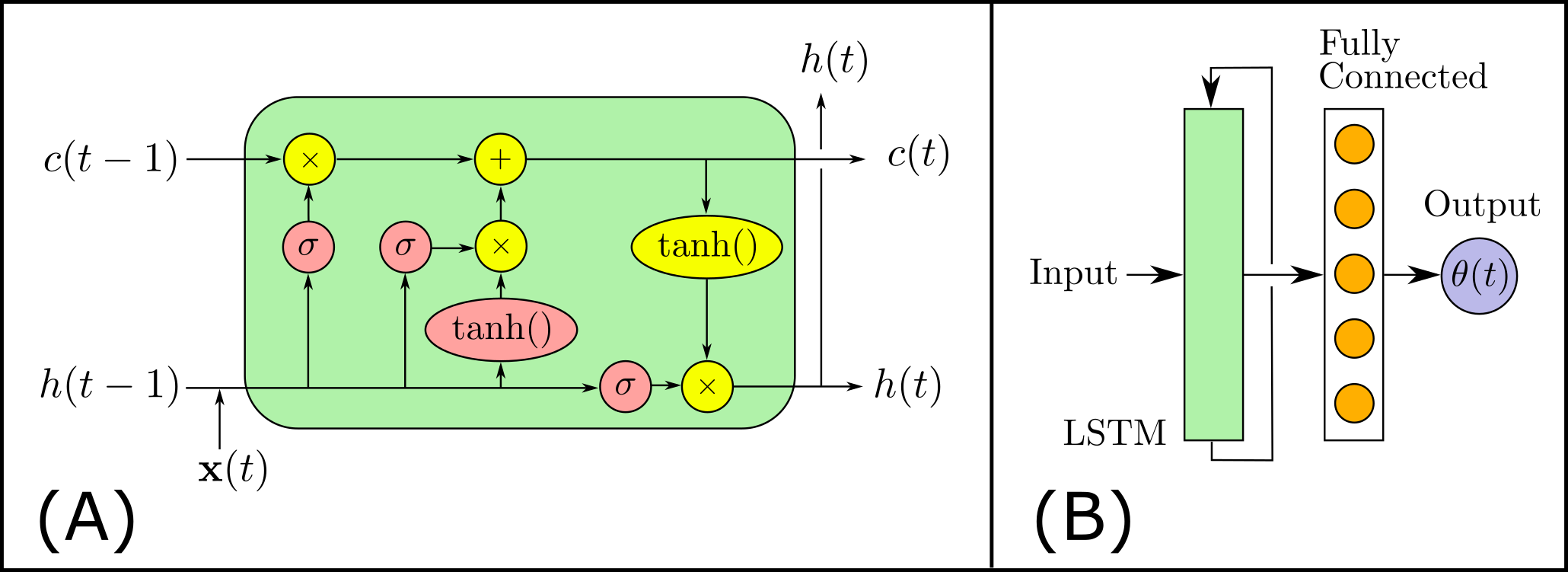}
    \caption{Neural network architecture. (A) Visualization of an LSTM block as represented by eqns. (\ref{eqn:first_lstm}-\ref{eqn:last_lstm}). (B) Our network consists of an LSTM layer connected to one fully-connected neuron layer.) }
    \label{fig:nn_structure}
\end{figure}

A single LSTM node (Fig. \ref{fig:nn_structure}(A)) takes an input $x(t) \in \mathbb{R}^n$, where $n$ is the number of input features, and outputs the cell vector $c_t \in \mathbb{R}$, given the inputs x(t), h(t-1), and c(t-1) with the following equations:

\begin{align}
f(t) &= \sigma(W_{f} x(t) + U_{f} h(t-1) + b_f) \label{eqn:first_lstm} \\
i(t) &= \sigma(W_{i} x(t) + U_{i} h(t-1) + b_i) \\
o(t) &= \sigma(W_{o} x(t) + U_{o} h(t) + b_o) \\
\tilde{c}(t) &= \tanh(W_{c} x(t) + U_{c} h(t-1) + b_c) \\
c(t) &= f(t) \circ c(t-1) + i(t) \circ \tilde{c}(t) \\
h(t) &= o(t) \circ tanh( c(t) ) \label{eqn:last_lstm}
\end{align}

\noindent where $\sigma$ is the logistic function, $\circ$ is component-wise multiplication, and the functions $f(\cdot), i(\cdot), o(\cdot) : \mathbb{R} \mapsto \mathbb{R}$ represent the forget gate, input gate, and output gate of the LSTM node, respectively. $W_{f,i,o,c} \in \mathbb{R}^{h \times n}$, $U_{f,i,o,c} \in \mathbb{R}^{h \times h}$, and $b_{f,i,o,c} \in \mathbb{R}^h$ are the matrices containing the respective input weight, recurrent weight, and bias parameters that are learned during training of the network, where $h$ is the number of hidden neurons.

Our neural network consists of three layers: one of LSTM nodes, one fully connected layer, and one output (Fig. \ref{fig:nn_structure}(B)).
Our model has 300 nodes in the LSTM layer and 300 nodes in the fully-connected neuron layer, which uses the ReLU activation function.
The fully-connected layer leads into a single node for the output actuation state: $\theta(t)$.
This approach is a simpler sequential network than the hierarchical LSTMs of \cite{luong_long_2021}, which was used for similar electrothermal twisted-coil polymer actuators.

\subsection{Test Setup and Training}

Our machine-learned model uses a time-window based representation for each configuration of sensor/actuation inputs ($v_1$ to $v_4$, eqns. \ref{eqn:v1}-\ref{eqn:v4}).
Each of these is supplemented with the previous deflection angle from our onboard sensor, as per time series dynamics:

\begin{equation}
X_k(t) = [v_k(t) \quad v_k(t-1) \quad \theta(t-1)]
\end{equation}

\noindent where $k \in{[1,2,3,4]}$ for each respective test using that input vector. 
Our predictor therefore models the dynamics function

\begin{equation}\label{eqn:thetahat}
\hat{\theta}(t) = g(X_k(t)).
\end{equation}

We used the mean-squared error (MSE) loss metric below to train the model and used the root-mean-squared-error metric (RMSE) to quantify the magnitude of the difference between our predictions for point $i$, $\hat \theta(t)_i,$ and the ground truth data $\theta(t)_i$. 

\begin{align}
MSE = \frac{1}{n}\Sigma_{i=1}^{n}\left( \hat \theta_i(t) - \theta_i(t) \right)^2, && 
RMSE = \sqrt{MSE} \label{eqn:mse}
\end{align}

\noindent The experimental data was split such that 67\% of the time series data was selected as training data while the remaining 33\% were chosen as the validation set. 
The bending angle data was linearly-scaled so that the maximum and minimum values lay between 1 and 0.
The models were trained using the Adam optimization algorithm with a learning rate of 10$^{-3}$ and weight decay of 0.9. 
Training was performed for 20 epochs, with convergence generally observed after 10 epochs for each model.
A batch size of 100 time steps was used; all other hyperparameters of the LSTM were unchanged from default values. 
The models implemented in Python 3.9 using the Keras API for Tensorflow running on a Google Colab notebook with a GPU-accelerated runtime environment, and has been open-sourced (see Data Availability Statement).

\subsubsection{One-Step Predictions vs. Open-Loop Rollouts}

Once these models were trained, we performed two experiments each to quantify their performance.
First, a single-step-ahead prediction in time for each of the input vectors was performed across the test set, i.e., eqn. (\ref{eqn:thetahat}) and eqn. (\ref{eqn:mse}) for all datapoints $i$ in the test set.
This is done to validate that these models have been sufficiently trained to  learning time-dependent information.
This test is useful primarily as a sanity check, since one-step predictions are only a basic minimum for model use (e.g. in predictive control).

Second, we employ our model in the form of its intended use: multi-step-ahead open loop rollout simulations.
Starting at some initial sample in the test set, indicated as $\widetilde{\theta}(t=0)$, our rollout recursion becomes

\begin{align}
\widetilde{X}_k(t) & = [v_k(t) \quad v_k(t-1) \quad \widetilde{\theta}(t-1)] \\
\widetilde{\theta}(t) & = g(\widetilde{X}_k(t)) \label{eqn:thetatilde},
\end{align}

\noindent i.e., predictions are fed back into the neural network for successive timesteps.
In this case, we used a measured value from the bend sensor as our initial value, but at all future points the model has no knowledge of the true deflection angle.

Finally, these rollout predictions were compared against a simple predictor as a reference for improvement gained by using our neural network.
We chose least squares regression for this task, which was performed on each test in the same manner with the same input vector $v_k$.
It is expected that the linear regression does not capture hysteresis, and therefore will not be able to predict the nuanced behaviors of our electrothermal actuators.

\section{Results}

\begin{figure}[h!]
    \centering
    \includegraphics[width=0.5\textwidth]{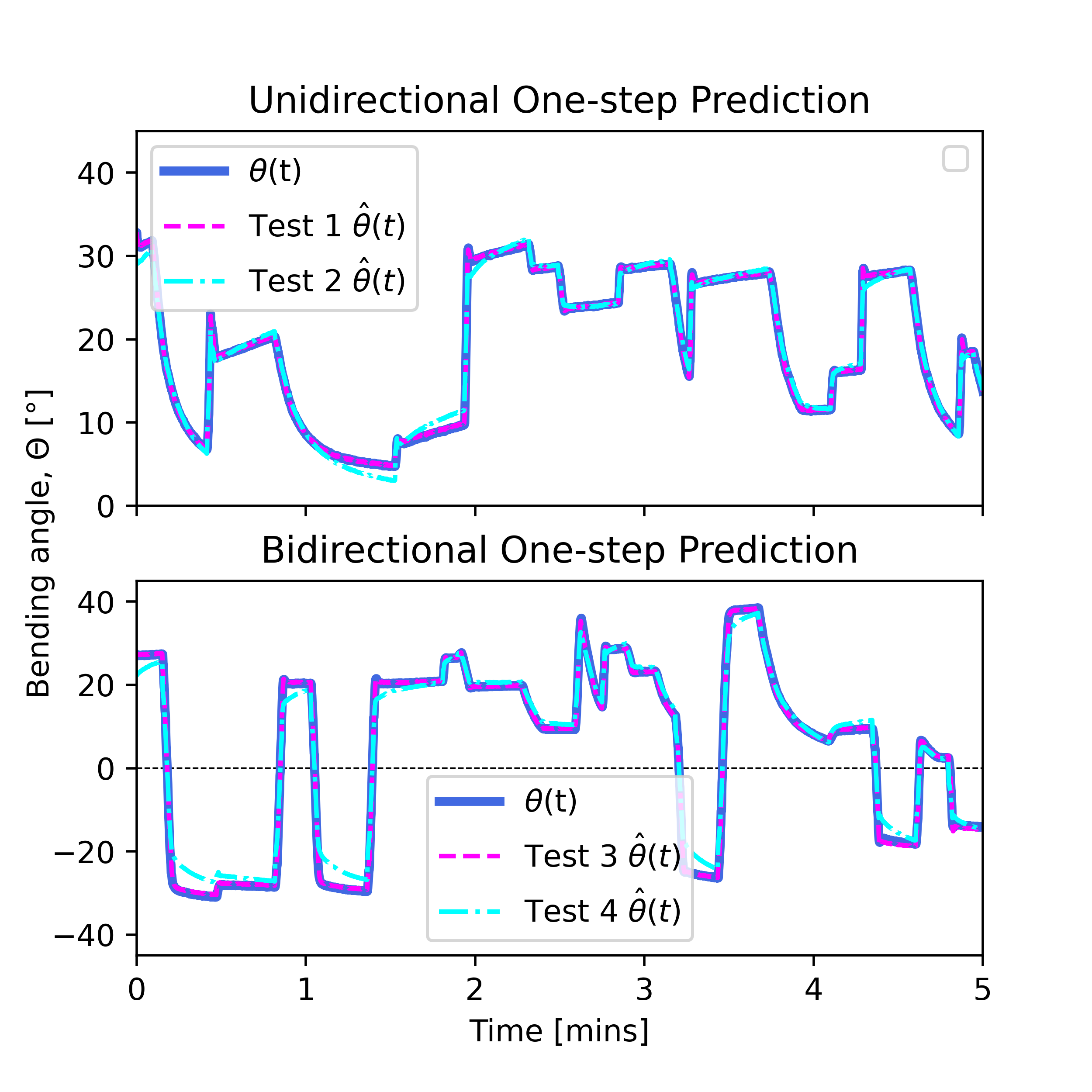}
    \caption{Predictions of each model one time-step into the future.
    (A) Models trained on unidirectional data using both PWM duty cycle and temperature as inputs, i.e., $v_{1,2}$. (B) Corresponding models trained on bidirectional data, $v_{3,4}$. Signals are virtually identical, indicating that one-step predictions are extremely accurate.} 
    \label{fig:onestep}
\end{figure}

Our framework was evaluated by examining the performance of our neural network predictor.
For the one-step predictions, $\hat{\theta}$ lies very close to the observed values $\theta$ in all cases (Fig. \ref{fig:onestep}). 
The best performing test in this case was Test $k=1$ for unidirectional bending with PWM duty cycle only (RMSE = 0.2091$^\circ$), while the worst performer was Test $k=4$ for bidirectional bending with PWM duty cycle and temperature as inputs (RMSE = 2.613$^\circ$).
In both cases, these tests were well within the manufacturer's stated accuracy of our soft bending sensor.

\begin{figure}[h!]
    \centering
    \includegraphics[width=0.95\textwidth]{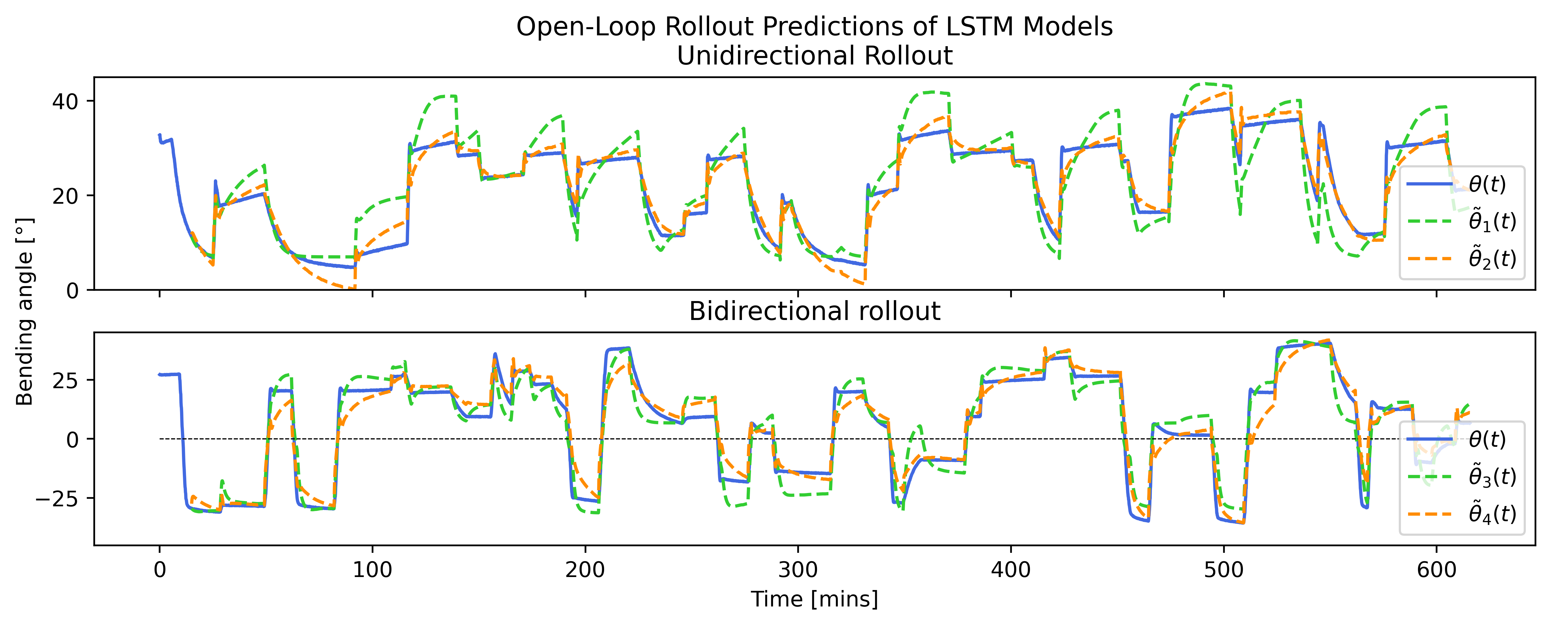}
    \caption{Predicted rollout trajectory over 10 minutes of actuation using $\theta$(t=15s) as initial value. (A) Test 1 depicts rollout from the model trained only on PWM input data on a single SMA. (B) Test 2 also depicts unidirectional rollout, but uses the model trained on both PWM and temperature. (C) and (D) show the rollout for the bidirectional models trained on PWM and on PWM with temperature, respectively. All tests show extremely little drift over arbitrarily long time scales.}
    \label{fig:rollouts}
\end{figure}

The results of open-loop rollout prediction across the various input spaces shows different tracking performance using different input spaces. 
Fig. \ref{fig:rollouts} compares the predicted trajectory of the limb over time using the models trained with each input vector $v_k$. 
In general, the fewer the number of input variables that were included, the worse the performance.
Ultimately, Test 4 performed the best of all the cases (RMSE = 5.350$^\circ$), while Test 1 performed the worst (RMSE = 6.296$^\circ$).
It is visually clear from Fig. (\ref{fig:rollouts}) that the bidirectional tests outperform the unidirectional tests, and that inclusion of temperature improves prediction quality.
However, all results demonstrate a usable model depending on the circumstance.

Our model also shows performance improvements over the naive least-squares fits.
Using PWM-only tests, least-squares predictions are effectively noise, whereas our model shows reasonably small error in this case (Table \ref{tab:performance}).
This is expected, since the LSTM network captures internal state.
Fits using temperature are lower error, primarily because temperature correlates reasonably well with bending angle.

\begin{figure}[h!]
    \centering
    \includegraphics[width=0.5\textwidth]{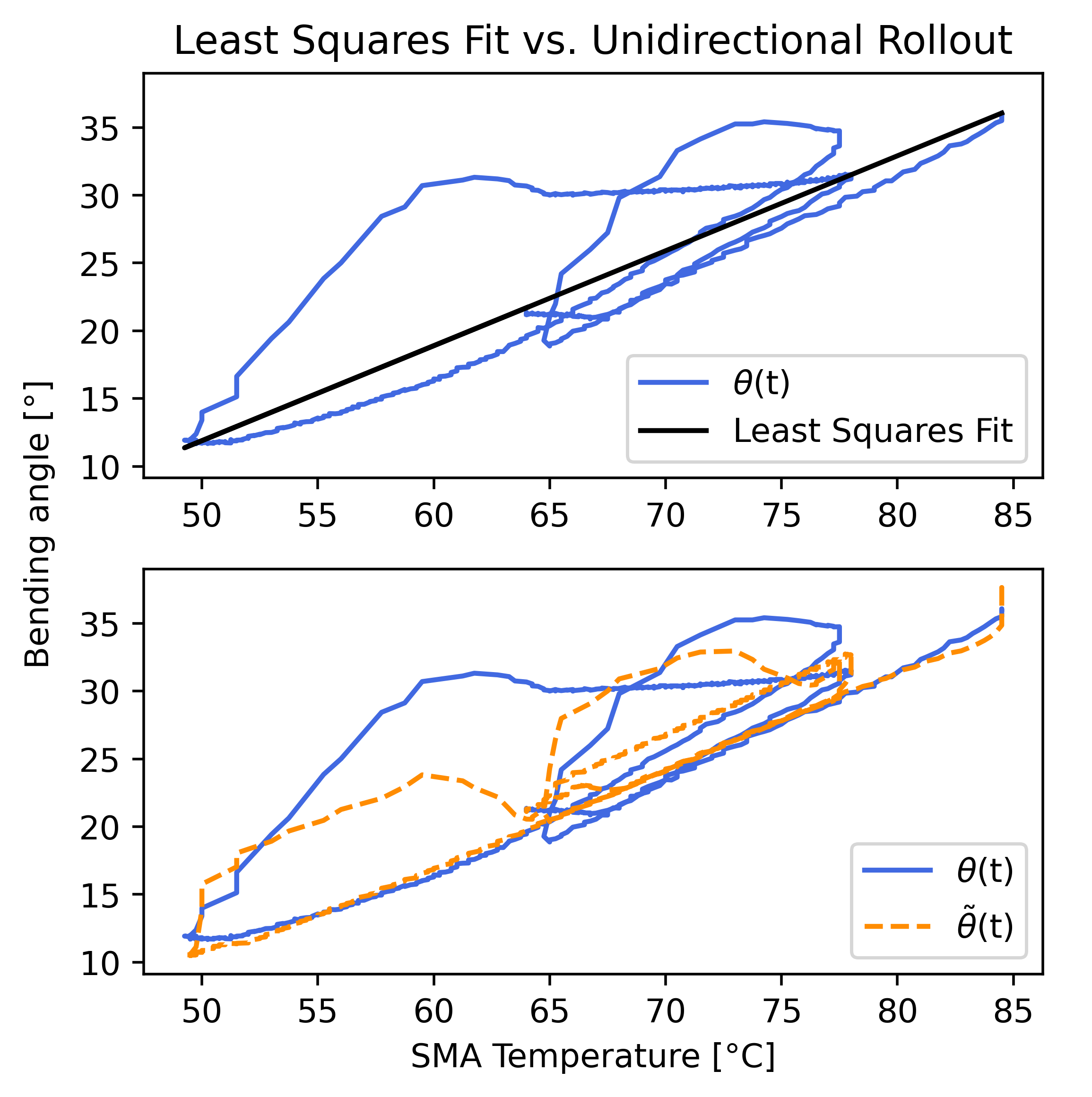}
    \caption{
    Demonstration of improved hysteresis characterization predicted from $\widetilde \theta$ rollout prediction on unidirectional bending trained on $u_1$ and $T_1$. The LSTM captures some hysteretic effects over time whereas a naive least-squares comparison does not.}
    \label{fig:hysteresis}
\end{figure}

However, the least-squares fit with temperature is artificially improved due to the density of our data.
Fig. (\ref{fig:hysteresis}) shows the least-squares fit for the unidirectional PWM-with-temperature test ($v_2$).
Since our SMAs spend more time cooling down than heating, the data set contains more cooling points.
The least-squares fit is biased toward this region, and provides a very poor prediction during heating, which is of higher importance for modeling and control.
In comparison, our LSTM neural network model attempts to account for hysteresis behavior during heating.

\section{DISCUSSION AND CONCLUSION}

This article introduces an embedded, \textit{in situ} sensor design and neural-network modeling framework for actuator dynamics in soft, SMA-powered robot limbs.
The sensing approach is inexpensive, off-the-shelf, and robust for extensive data collection.
The learned LSTM-based neural network dynamics models appear to capture the core physics that govern these electrothermally-actuated soft limbs.

\subsection{Neural Network Model Performance Analysis}

Single-step-ahead predictions validate that our model sufficiently learned the importance of time-dependent behaviors, and open-loop rollouts produce low-drift predictions over extremely long time periods (i.e. $\sim$10 minutes).
In the scenario of the most amount of data -- bidirectional bending with temperature included -- predictions are of low error in all regions of the state space.
The RSME error in the resultant predictions were on the same order of our sensor accuracy.
Predictions were performed 2.4$\times$ faster than real-time, which is promising for feedback control applications.
However, improvements can still be made to reduce error.

When training on only the PWM duty cycles, particularly in the unidirectional case (Fig. \ref{fig:rollouts}(A)), our model consistently overshoots the expected deflection angle. 
This could be attributed to the availability of less data due to fewer inputs, or more mechanically hysteretic behaviors with single-SMA motions -- i.e., the only elastic restoring force is due to the soft silicone.  
However, in the bidirectional case, rollouts using PWM inputs only show surprisingly accurate predictions.
As a result, it may be possible to model these electrothermally-actuated soft robot limbs without temperature sensing.
Mechanical designs for untethered soft thermal actuator robots can therefore be made simpler while still maintaining our capability for dynamics predictions, and different components of our framework can be used independently.
Possible future work includes experiments with more neural network layers and architectures, as well as tuning of hyperparameters.

Inclusion of temperature is observed to reduce or eliminate overshoot in the predictions.
The RMSE error improvement is small over the course of the entire test set in the bidirectional case (approximately 13\% lower error); however, visual inspection of Fig. (\ref{fig:rollouts}) shows qualitative improvements.

\subsection{Sensor Design Implications}

Our temperature sensor remained attached and showed no visible fracture or detachment even after our extensive testing.
However, the proof-of-concept fabrication process used here, where thermally conductive epoxy is applied by hand, often led to large amounts of adhesive and therefore a large thermal mass at the junction site.
Correspondingly, the neural network's response in predicted angle has a slight time delay in comparison to the test data.
Many of the results in Fig. (\ref{fig:rollouts}) do not appear to capture sharp increases in bending angle.

\subsection{Future Work and Conclusion}

This work contributes an approach for embedded, in-situ sensing and dynamics predictions of electrothermally-actuated soft robot limbs.
Our proof-of-concept demonstrates accurate predictions on the order of our soft deflection sensor's capabilities in a compact device architecture that can be readily translated to untethered soft robots.
The framework therefore bridges a crucial gap between hysteretic, difficult-to-model actuators and practical use in soft robotics applications.

Future directions for this work include improvements to both the sensor and the model, as well as application in walking robots.
For the sensor, reduced thermal mass may be possible with a flexible adhesive as opposed to a rigid epoxy.
We anticipate examining materials such as the liquid metal embedded elastomers in \cite{bartlett_high_2017} and \cite{malakooti2019liquid}, which exhibit high thermal conductivity, mechanical compliance, and fracture toughness (\cite{kazem2018extreme}).
The manufacturing process may also be automated using digital fabrication techniques like 3D printing to reduce human error.

For the predictive model, more tailored neural network architectures may be explored, such as those in \cite{kang_learning-based_2020,han_use_2018}.
Approaches such as the Preisach model (\cite{ge_preisach-model-based_2021}), which have seen success in system identification for SMAs, could also be used.
We anticipate implementing our method on a walking robot and using those results to quantify the need for model improvement before exploring algorithms beyond our proof-of-concept.

Most importantly, future work can incorporate these models as part of predictive control or planning algorithms for positioning of untethered soft limbs.
Whether robot designs include the proposed temperature sensor, or simply use PWM duty cycle in the neural network, our results imply that predictions are possible.
Robot designs are under investigation that incorporate our framework as part of a walking soft robot.

Lastly, we note that the dynamics model tests presented here utilize ground-truth temperature measurements during open-loop rollouts.
These would not be available during online prediction scenarios, as would be the case for feedback control.
However, we consider these tests to be valid, since prior work has shown that the temperature of SMA wires can be predicted reasonably well from Joule heating models (\cite{wertz_trajectory_2022}) and that LSTM networks may also be able to predict temperature itself (\cite{luong_long_2021}).
A possible direction for future work is therefore predicting temperature alongside bending angle.
Our temperature-free models, which do not have this drawback, can still be used for true open-loop predictions.


\section*{Conflict of Interest Statement}

The authors declare that the research was conducted in the absence of any commercial or financial relationships that could be construed as a potential conflict of interest.

\section*{Author Contributions}

AS developed the concept and design of the study, as well as the hardware platform design, fabrication procedure, and prototype.
RM developed the neural network architecture and software, and performed the simulations.
AW developed the electronics and embedded software for the hardware prototype.
AS and RM analyzed the data and wrote the initial draft of this manuscript.
CM provided editing and feedback as well as study direction.
All authors contributed to manuscript revision, read, and approved the submitted version.

\section*{Funding}

This work was supported by an Intelligence Community Postdoctoral Research Fellowship through the Oak Ridge Institute for Science and Education and the National Science Foundation (NSF) National Robotics Initiative (Award: 1830362).

\section*{Acknowledgments}

The authors would like to thank all members of the Soft Machines Lab at Carnegie Mellon University for their helpful discussions and feedback on this article.
In particular, we acknowledge Zach J. Patterson for collaboration in the concept for the sensor used here.

\section*{Data Availability Statement}
The datasets generated and analyzed for this study can be found in the following repository on Github: \href{https://github.com/rkme/memalloy}{https://github.com/rkme/memalloy}.

\bibliographystyle{Frontiers-Harvard} 
\bibliography{sources}

\section*{Figure captions}

{\renewcommand{\arraystretch}{1.2}
\begin{table}[h!]
\centering
\caption{Comparison of RMSE of least squares fit vs. learned model rollout, bidirectional motions}
\begin{tabularx}{1\columnwidth}{ |l|l|X| }
 \hline
 Test Case & Least Squares RMSE & LSTM Rollout RMSE\\ 
 \hline
 Test 3 ($u_j$ only) & 19.01$^\circ$ & 6.151$^\circ$ \\ 
 Test 4 ($u_j$ and $T_j$) & 6.372$^\circ$  & 5.350$^\circ$ \\
 \hline
\end{tabularx}
\label{tab:performance}
\end{table}
}

\end{document}